\documentclass[conference]{IEEEtran}
\IEEEoverridecommandlockouts
\usepackage{cite}
\usepackage{url}
\usepackage{multirow}
\usepackage{amsmath,amssymb,amsfonts}
\usepackage{algorithmic}
\usepackage{graphicx}
\usepackage{textcomp}
\usepackage{xcolor}
\def\BibTeX{{\rm B\kern-.05em{\sc i\kern-.025em b}\kern-.08em
    T\kern-.1667em\lower.7ex\hbox{E}\kern-.125emX}}
\begin{document}

\title{A Preliminary Analysis of Automatic Word and Syllable Prominence Detection in Non-Native Speech With Text-to-Speech Prosody Embeddings
}

\author{\IEEEauthorblockN{Anindita Mondal, Rangavajjala Sankara Bharadwaj, Jhansi Mallela, Anil Kumar Vuppala, Chiranjeevi Yarra}
\IEEEauthorblockA{\textit{Language Technologies Research Center,}{ IIIT Hyderabad, India.}
}
}

\maketitle

\begin{abstract}
Automatic detection of prominence at the word and syllable-levels is critical for building computer-assisted language learning systems. It has been shown that prosody embeddings learned by the current state-of-the-art (SOTA) text-to-speech (TTS) systems could generate word- and syllable-level prominence in the synthesized speech as natural as in native speech. To understand the effectiveness of prosody embeddings from TTS for prominence detection under nonnative context, a comparative analysis is conducted on the embeddings extracted from native and non-native speech considering the prominence-related embeddings: duration, energy, and pitch from a SOTA TTS named FastSpeech2. These embeddings are extracted under two conditions considering: 1) only text, 2) both speech and text. For the first condition, the embeddings are extracted directly from the TTS inference mode, whereas for the second condition, we propose to extract from the TTS under training mode. Experiments are conducted on native speech corpus: Tatoeba, and non-native speech corpus: ISLE. For experimentation, word-level prominence locations are manually annotated for both corpora. The highest relative improvement on word \& syllable-level prominence detection accuracies with the TTS embeddings are found to be 13.7\% \& 5.9\% and 16.2\% \& 6.9\% compared to those with the heuristic-based features and self-supervised Wav2Vec-2.0 representations, respectively. 
\end{abstract}

\begin{IEEEkeywords}
FastSpeech2, prosody embeddings, prominence, TTS, native speech, non-native speech
\end{IEEEkeywords}

\section{Introduction} \label{intro}
\vspace{-0.1cm}

Speech serves as a multilayered signal, carrying not only the essence of lexical and grammatical content but also the nuances of prosody and distinctive features tied to individual speakers (such as identity and emotional nuances). Prosody, which extends beyond individual phonetic units, covers elements related to syllables, words, phrases, sentences, and more extended utterances, collectively known as supra-segmental features \cite{werner, lehiste1970suprasegmentals}. Prominence (also referred to as stress), within this context, manifests itself as a greater emphasis on the syllable or words \cite{cutler2005lexical}. The acoustic correlation of the emphasis can be observed through the manifestation of duration, energy, and pitch. Native speakers acquire their language naturally from early childhood, whereas nonnative speakers, often influenced by their native language phonological aspects, may introduce distinctive stress patterns when speaking the second (L2) language. Investigating stress across different linguistic levels is pivotal for unraveling its intricate implications, drawing substantial attention in the realm of research, particularly in computer-assisted language learning (CALL) and non-native accented speech synthesis. A multitude of studies has scrutinized stress, probing into its manifestations at word-level \cite{wp1,wp3, wp2} as well as syllable-level \cite{Tepperman, yarra19b_slate, 7953277, vae-dnn}. An unsupervised approach for automatic word prominence detection is presented in \cite{wp1} considers spectral and temporal features along with word-level part-of-speech (PoS) tags. Sonority, which represents the carrying power of sounds in words or longer utterances, plays a key role in the method proposed by \cite{7953277} for automatic syllable stress detection. Most of the works in the literature \cite{ wp1,wp3, wp2,Tepperman, yarra19b_slate, 7953277, vae-dnn} explored the heuristics based features and a very few works considered the self-supervised embeddings \cite{NCC_W2VEC}. Generally, these embeddings were proposed to represent primarily phonemic information than prosodic. On the other hand, current state-of-the-art text-to-speech (TTS) models were proposed to learn the representations specific to prosodic information. To the best of our knowledge, no work considers prosody representations from TTS for word and syllable-level prominence detection task. \par 
Recent advancements in TTS research have prominently focused on incorporating expressiveness, prosody, and achieving more natural speech synthesis. Tacotron2 \cite{ Tacotron2} is a popular TTS, which is an improved version of Tacotron \cite{Tacotron}, that generates good quality speech. Glow-TTS \cite{ Glow-TTS} utilizes a generative model based on Glow, a normalizing flow model \cite{Glow} and Variational Inference Text-to-Speech (VITS) \cite{vits} leverages variational inference \cite{flow} for efficient and high-quality text-to-speech synthesis. FastSpeech2 \cite{ren2022fastspeech}, improves upon its predecessor, FastSpeech \cite{FastSpeech}, and effectively incorporated the prosodic information in its synthesized speech. This is done by tackling the one-to-many mapping using its variance adaptor that effectively generates prosodic variations such as pitch, energy and duration.
\par 
In this study, the aim is to explore prosody embeddings from FastSpeech2 for word and syllable-level prominence detection using ``text-only" and ``speech plus text". Further, their effectiveness is compared with the existing prominence detection methods considering both native and non-native speech. We consider the energy, duration and pitch embeddings from Fastspeech2 for our analysis using native and non-native speech corpora (Tatoeba \cite{tatoeba_url} and ISLE \cite{ISLE}). A selected subset of 3000 audios from these two datasets were annotated with word-level prominence and the comparative analysis is performed in the following three stages: 1) with Principal Component Analysis (PCA), 2) computing similarity and dissimilarity measures between features of stressed and unstressed groups and 3) classification based analysis in supervised (DNN Classifier) and unsupervised (K-Means clustering) manner. The study reveals that the embeddings obtained under ``speech plus text" case show better discrimination than those obtained under ``text-only" case in  native and non-native speech. Under non-native speech, the highest relative improvement on \textbf{word \& syllable-level} prominence detection accuracies with the TTS embeddings are found to be \textbf{13.7\% \& 5.9\%} and \textbf{16.2\% \& 6.9\%} compared to those with the heuristics-based features and self-supervised Wav2Vec-2.0 representations respectively. Further, it is found that the embeddings provide superior performance for German as compared to Italian speakers.

\begin{figure*}[t]
  \centering
  \includegraphics[width=0.85\textwidth, height=5.7cm]{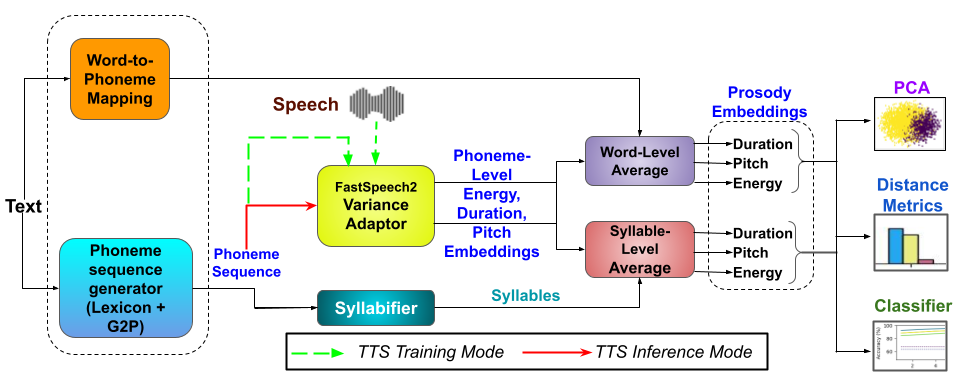} 
  \vspace{-0.3cm}
  \caption{Block Diagram showing the prosposed approach for obtaining embeddings from Fastspeech2 variance adaptor}
  \label{fig:blockdiagram}
  \vspace{-0.6cm}
\end{figure*}

\section{Motivation} \label{mv}
\vspace{-0.1cm}
There have been various methods designed to model word and syllable prominence in speech as elaborated in \ref{intro}. These methods extract complex features from the speech signal and utilize them in a manner that aids in stress detection. In our work, we are focusing on exploring the intermediate representations learned by TTS systems and understanding how these contribute to prominence detection. By investigating these intermediate prosody representations, we aim to uncover how TTS systems capture and utilize prosodic features. This understanding can provide valuable insights into enhancing TTS models, making them more effective in generating natural and intelligible speech across different languages and speakers.\par
The motivation behind selecting intermediate representations lies in the unique capabilities of FastSpeech2, particularly its variance adaptor. During training, the variance adaptor learns to extract phoneme-level duration, pitch, and energy embeddings from the input speech and text. These embeddings capture crucial prosodic characteristics of speech, such as the timing of phonemes, pitch variations, and energy levels. During inference, the model utilizes these learned embeddings to generate speech with appropriate intonation, stress, and rhythm, ensuring the naturalness and expressiveness of the synthesized utterances. Our interest lies in exploring the efficiency of these embeddings in modeling prominence in both native and non-native speeches. By investigating these intermediate embeddings, we aim to understand how effectively the model grasps subtleties in duration, energy and pitch as these are crucial for conveying prominence in speech. \par
Exploring intermediate prosody embeddings in speech synthesis models like FastSpeech2 becomes particularly motivating when considering the challenge of generating non-native speech with appropriate linguistic variations using a TTS system trained primarily on native speech data. This difficulty arises from the inherent differences in prosodic patterns, phonological structures, and cultural nuances between native and non-native speakers. Understanding how these models capture and utilize prosodic features can assess their ability to adapt to both native and non-native speech patterns. This exploration holds promise for catering to a wide range of speakers and language learners.
\section{Dataset} \label{Dataset}
\vspace{-0.1cm}

   \textbf{Tatoeba} \cite{tatoeba_url} stands as an expansive and openly accessible compilation of English sentences accompanied by high-quality translations spanning over 300 languages. It serves as a comprehensive repository facilitating linguistic exploration and multilingual understanding \cite{tatoeba,Tiedemann2012ParallelDT}. This dataset is continuously expanding through voluntary contributions from numerous members.
We curated a subset of the Tatoeba Dataset, specifically focusing on English audio files accompanied by their transcripts. Each audio file within this subset underwent a thorough manual annotation process to accurately identify and label prominence. This subset encompasses approximately 7122 instances of English words, excluding silences and noise. Adopting a word-centric unit of analysis, we successfully transcribed the prominence markers to the word-level, ensuring precise identification, and consolidated the prominence labels into a binary classification: 1 for prominent words and 0 for non-prominent words. This annotation resulted in 1083 prominent words and 6039 non-prominent words. \par
The \textbf{ISLE Corpus} \cite{ISLE} comprises 7,834 speech utterances from 46 non-native English learners (23 are German (GER), and (23 are Italian (ITA)). NIST \cite{NIST} syllabification software is used to derive syllable transcriptions, and aligned syllable boundaries are obtained from phone transcriptions. It had 48868 syllables as stressed and 16693 syllables as unstressed. A combined subset of 2000 utterances, with 1000 each from German and Italian speakers, underwent word-level annotation, mirroring the methodology used in the Tatoeba dataset. We specifically opted for a dataset from non-native speakers to thoroughly analyze the variability of stress patterns in both syllables and words. 

\section{Proposed Methodology} \label{ method}
\vspace{-0.1cm}
Figure \ref{fig:blockdiagram} shows a block diagram describing the steps involved in embedding extraction, which is developed based on the variance adaptor in FastSpeech2. The block diagram has the following steps: 1) mapping from word sequence to phoneme sequence, 2) the variance adaptor in FastSpeech2, 3) syllabifier for converting phoneme sequence to syllable sequence, and 4) word and syllable-level averaging for deducing the embeddings for the respective words and syllables.
\subsection{FastSpeech2 Variance Adaptor} \label{VarianceAdaptor}
\vspace{-0.1cm}
The variance adaptor is a neural network framework specifically designed to extract embeddings, which correspond to the following three acoustic components associated with the prosody: duration, energy and pitch. In the training phase, ground truth duration, pitch, and energy values extracted from the speech are employed as targets to train the corresponding predictors. Using the three acoustic components extracted from the speech, it learns the association between the text and embeddings. Thus the training requires parallel speech and text data, referred to as ``speech plus text" case. During the inference stage, with the learned association, the variance adaptor generates embeddings representing the three types of acoustic components using only text, referred to as ``text-only" case. In both stages, the embeddings are computed for each phoneme in the phoneme sequence of the text, which is obtained from an inbuilt word-to-phoneme mapping tool based on pronunciation lexicon and grapheme-to-phoneme (G2P) conversion. A user is facilitated to modify the entries in the lexicon according to the requirement. The variance adaptor is a 2-layer 1D-convolutional network with ReLU activation, layer normalization, dropout, and a linear layer, using a kernel size of 3 and an output sequence size of 256.
\par  
\emph{``Speech plus text" case vs ``text-only" case:} The overall process of extracting embeddings remains consistent for both ``text-only" and ``speech plus text" cases, with the exception of the dotted arrow depicted in the block diagram. This dotted arrow is exclusive to the embedding extraction process during the ``speech plus text" case, where actual duration, pitch, and energy values extracted from the speech serve as input to the FastSpeech2 variance adaptor operating under the training mode. However, for the ``text-only" case, the variance adaptor generates the embeddings in the inference mode solely from the text without speech. In this work, the aim is to analyse the embeddings from both cases as the prosody in the speech can vary for a given text. Thus, in the ``text-only" case, one-to-many possibilities exist, whereas. in the ``speech plus text" case, the obtained prosody embeddings could capture prosody in the speech due to the usage of the speech in the training process. Therefore, the anticipation is that the ``speech plus text" approach offers more comprehensive prosody embeddings compared to the ``text-only" case, which solely relies on textual cues. This motivated to analyse the embeddings extracted in these two distinct modes.

\subsection{Word \& syllable-level embeddings extraction}
\vspace{-0.1cm}
The variance adaptor provides the embeddings for each phoneme in the phoneme sequence obtained from the text after the word-to-phoneme mapping step.\par 
\emph{``Text-only" case:} The word and syllable-level embeddings are obtained from the phoneme level embeddings for each phoneme associated with the text. For this, phoneme level embeddings are averaged across all the phonemes within each syllable or a word associated with the text, respectively to obtain syllable and word-level embeddings. The embeddings derived from variance adaptor exhibit dimensions of (number of phonemes $ \times$  $256$) and are subsequently transformed to dimensions of (number of words $ \times$  $256$) and (number of syllables $ \times$  $256$), respectively, for word and syllable-level embeddings. The phonemes within each word are mapped using inbuilt word-to-phoneme mapping in the variance adaptor. Whereas, the phonemes association with syllables are obtained with a syllabifier, which maps one or more phonemes to a syllable thereby phoneme sequence associated with the text is converted to a respective syllable sequence.

\begin{figure*}[h]
  \centering
  \includegraphics[width=1\textwidth, height = 4 cm]{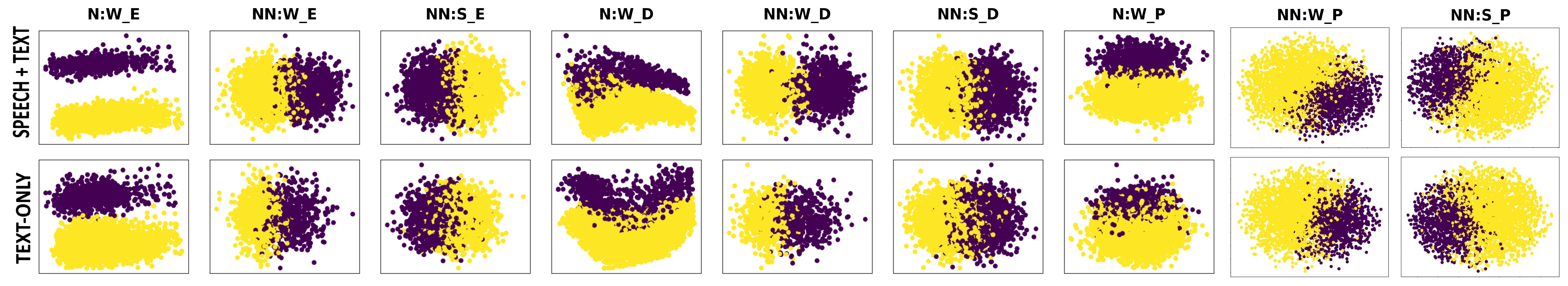}
  \vspace{-0.7cm}
  \caption{Scatterplots comparing two principal components under Speech+Text and Text-only cases.('N' denotes native, 'NN' denotes Non-native, 'W': word-level, 'S': syllable-level, 'E': Energy, 'D': Duration and 'P':Pitch)}
  \label{fig:pca1}
  \vspace{-0.6cm}
\end{figure*}

\emph{``Speech plus text" case:} Variance adaptor uses the phoneme sequence obtained from the inbuilt lexicon, which is based on native English speech. Typically, the lexicon contains multiple pronunciations and one of these is selected with the Montreal force-aligner used in the variance adaptor during the training. This process may result in processing errors in picking the correct pronunciation that exists in the speech. These errors would be more in the non-native speech i.e., English speech from German and Italian speakers in the ISLE corpus. To overcome these errors, the manual phonetic transcriptions available in the corpus are utilized by modifying the pronunciation lexicon to incorporate phoneme sequences for each word based on the manual transcriptions. This modification is performed dynamically for each speech and text pair while extracting the embeddings. The variance adaptor, while running in the training mode, refers to this modified lexicon for the embedding extraction. The training mode is run for multiple epochs by fine-tuning the pre trained LJSpeech \cite{ljspeech17} model. We study the training for every epoch to assess whether the embeddings produced with increased learning iterations enhance prominence detection as the model becomes more proficient over time. 
\section{Experimental Setup} \label{baseline}
\vspace{-0.1cm}
In our approach, we obtain both word-level and syllable-level embeddings for duration, pitch and energy using ISLE Corpus, by leveraging two distinct cases: ``speech plus text" and ``text-only" . However, for Tatoeba corpus, we extract only word-level embeddings for both the cases as the syllable-level ground-truth markings are not available. Manual annotation of those requires expertise, which is costly and time consuming. Hence, we could not perform syllable-level prominence analysis on the Tatoeba data. The prominence analysis at both the word and syllable-levels is performed in three stages:
\begin{enumerate}
    \item Using Principal Component Analysis (PCA) as it allows us to visualize linear projected nuanced variations in embeddings at both the word and syllable-levels
    \item Computing the distances between stressed and unstressed words or syllables.
    \item Considering supervised (DNN classifier) and unsupervised (K-Means clustering) approaches.
\end{enumerate}
The feature computation and experimental setup for detecting word and syllable-level prominence are described below.

\emph{{Word Prominence:}} Following the work reported in \cite{wp3}, prosodic information such as area under the F0 curve, Voiced-to-unvoiced ratio, F0 peak/valley amplitude and location etc and lexico-syntactic information like part-of-speech tags, word type (content word or a function word) were used to obtain \textbf{heuristics-based features} from the word-level annotated subset of audios in the ISLE and Tatoeba datasets. These features were used to train a  deep neural network (DNN) model having 6 dense layers along with Batch Normalization and Dropout in its architecture. After the first Dense layer (64 units) and the third Dense layer (32 units),  Rectified Linear Unit (ReLU) activation function is applied, followed by Batch Normalization and Dropout (0.3). This DNN model underwent training for 50 epochs with Adam optimizer and binary cross-entropy loss function to perform word-level stress detection. \par
\emph{{Syllable Prominence:}} We utilize the state-of-the-art 19-dimensional acoustic-based features, accompanied by 19-dimensional binary features capturing context dependencies (hereafter referred to as \textbf{heuristics-based features}), in line with \cite{yarra19b_slate}. Following the approach in \cite{vae-dnn}, a DNN classifier, comprising 8 hidden layers with ReLU activation, Adam optimizer, and binary cross-entropy loss function, is trained for 200 epochs to perform syllable-level stress detection. \par
\emph{{Wav2Vec-2.0 Representations:}} Building on the approach in \cite{NCC_W2VEC}, self-supervised Wav2Vec-2.0\cite{wav2vec} representations are first extracted at the frame level to capture detailed acoustic information across time. These frame-level features are then averaged to align with word and syllable boundaries, enabling an effective representation of non-native speech patterns in the ISLE corpus for further analysis.  
\par
The heuristics features, Wav2Vec-2.0 representations and the three types of TTS based embeddings are utilized in distance and classification based analysis whereas, we perform PCA analysis only on TTS based embeddings. The recorded accuracies for DNN classifiers and K-means clusters with the heuristics-based features and Wav2Vec-2.0 representations serve as a baseline to assess the effectiveness of the extracted TTS based embeddings.
\section{Results}
\vspace{-0.1cm}
\subsection{PCA Based Analysis}
Figure \ref{fig:pca1} shows the scatterplots of the two principal components obtained from the energy, duration and pitch embeddings under ``speech plus text" and ``text-only" cases at word and syllable-levels under native and non-native speech conditions. The observations from the figure indicate that there is less overlap between stressed (purple dots) and unstressed (yellow dots) elements in energy embeddings as compared to duration and pitch embeddings, for both ``speech plus text" and ``text-only" cases at the word and syllable-levels. This indicates that the energy embeddings offer superior separation compared to duration and pitch embeddings. Notably, the embeddings exhibit less overlap when considering ``speech plus text" compared to ``text-only" cases. The findings suggest that incorporating speech data enhances the quality of the embeddings for prominence detection and this aligns with the claim mentioned in section \ref{VarianceAdaptor} that prominence is particularly present in speech and to some extent in text.

\vspace{-0.1cm}
\subsection{Distance Metrics}
Drawing inspiration from \cite{distance} that distinctly delves into proximity measures tailored for numeric data attributes, the seven metrics were computed as listed below:

\begin{enumerate}
\item \textbf{Similarity Measures:}
\begin{enumerate}
    \item
Cosine Similarity \( =\frac{\sum_{i=1}^{n} x_i \cdot y_i}{\sqrt{\sum_{i=1}^{n} x_i^2} \cdot \sqrt{\sum_{i=1}^{n} y_i^2}}\)
\vspace{0.12cm}
    \item
Jaccard Similarity \(= \frac{|X \cap Y|}{|X \cup Y|}\)
\end{enumerate}

    \item  \textbf{Dissimilarity Measures:}
        \begin{enumerate}
            \item 
        Manhattan Distance \( = \sum_{i=1}^{n} |x_i - y_i|\)
        \vspace{0.12cm}
            \item
        Euclidean Distance \(= \sqrt{\sum_{i=1}^{n} (x_i - y_i)^2}\)
        \vspace{0.12cm}
            \item
        Chebyshev Distance \( = \max_{i}(|x_i - y_i|)\)
        \vspace{0.12cm}
            \item
        Canberra Distance \(= \sum_{i=1}^{n} \frac{|x_i - y_i|}{|x_i| + |y_i|}\)
        \vspace{0.12cm}
            \item
        Mahalanobis Distance \\ \( = \sqrt{(x_i - y_i)^T \cdot S^{-1} \cdot (x_i - y_i)}\) where \(S\) is the covariance matrix. 
        
        \end{enumerate}
\end{enumerate}
\par

In the above measures, \(x_i \in X\) and \(y_i \in Y\), where \(X\) and \(Y\) denote the subset of features that belong to stressed and unstressed categories, respectively. In this work, the subset of features are taken from the following five types: 1) energy embeddings, 2) duration embeddings, 3) pitch embeddings, 4) heuristics-based features and 5) Wav2Vec-2.0 representations. Cosine and Jaccard capture resemblance, while the others capture dissimilarity, so stressed and unstressed features should show lower similarity and higher dissimilarity values for better discrimination. \par

\begin{figure}[h]
  \centering
  \includegraphics [width=0.5\textwidth, height =6cm] {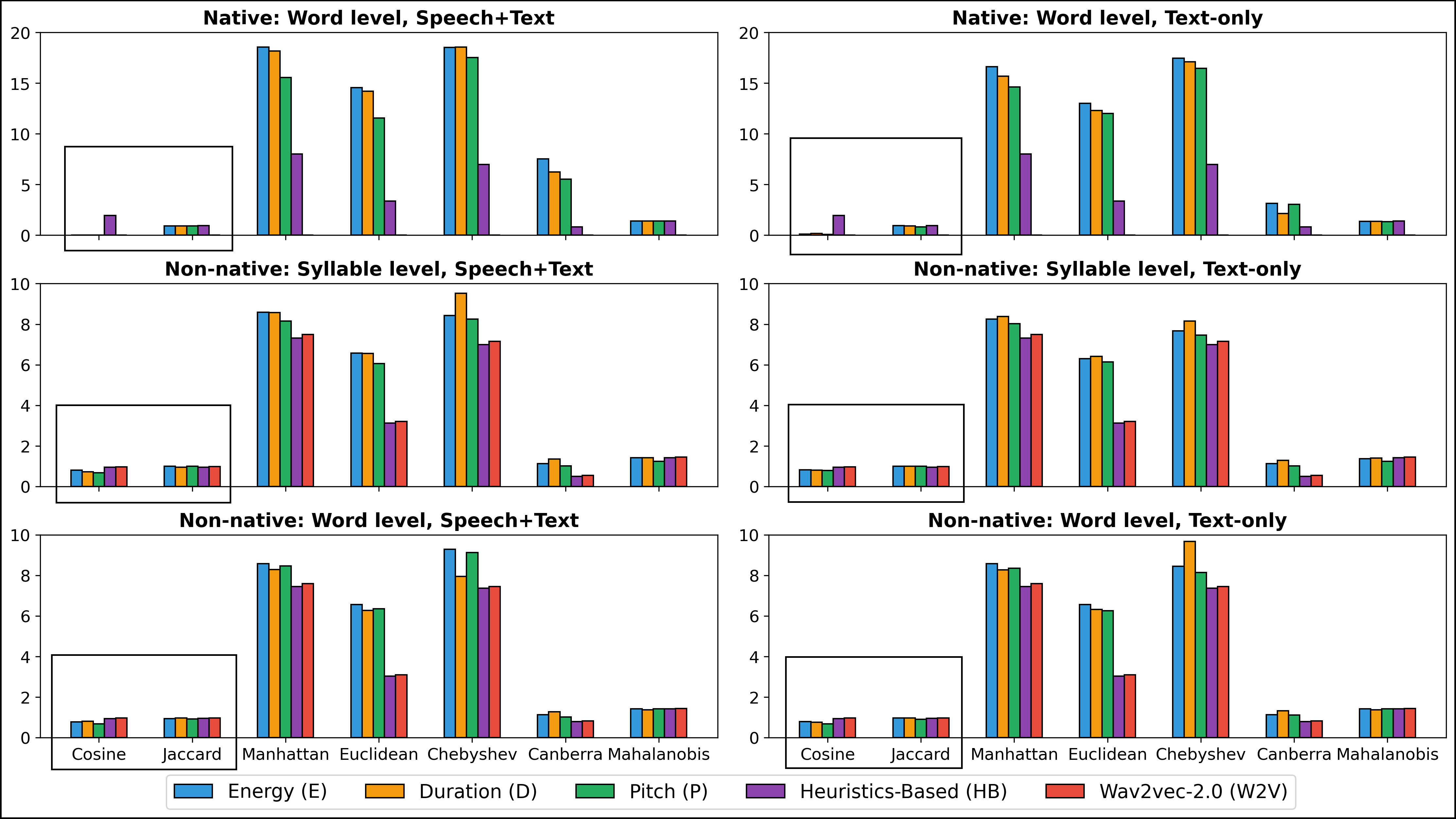}
  \caption{Comparison of Distance metrics for native and non-native speech under Text-only and Speech+Text cases (the distances enclosed within the box indicate similarity measures)}
  \label{fig:distances}
\end{figure}

The seven metrics, computed between stressed and unstressed words and syllables, employing the five distinct feature sets, are shown with bar graph in Figure \ref{fig:distances}. It is observed that the dissimilarity measures of prosody embeddings exhibit greater values compared to the heuristics and Wav2Vec-2.0 features and is further supported by the lower values of the similarity measures, indicating a higher degree of dissimilarity between the corresponding representations. The distinct patterns in the dissimilarity measures highlight the unique characteristics and separability of pitch, energy and duration compared to heuristics-based features and Wav2Vec-2.0 representations between stressed and unstressed words and syllables. 

\vspace{-0.1cm}
\subsection{Classification Analysis}
Table \ref{table:Accuracies} shows the classification accuracies at word and syllable-level with K-Means and DNN considering all three types of embeddings in comparison with heuristics-based features and Wav2Vec-2.0 representations. 
\par

\begin{table}[ht]
\setlength{\tabcolsep}{3pt}
    \caption{Comparison of Accuracies for Word and Syllable-Level Prominence Detection (K-M denotes K-Means, E, D, P, HB and W2V denote Energy, Duration, Pitch embeddings, Heuristics-based and Wav2Vec-2.0 Features respectively)}
\centering
\begin{tabular}{cccccccccc}
\cline{3-10}
                                                                                                      & \multicolumn{1}{c|}{}             & \multicolumn{4}{c|}{\textbf{Word Prominence}}                                                                                                   & \multicolumn{4}{c|}{\textbf{Syllable   Prominence}}                                                                                             \\ \cline{3-10} 
                                                                                                      & \multicolumn{1}{c|}{}             & \multicolumn{2}{c|}{\textbf{Speech + text}}                             & \multicolumn{2}{c|}{\textbf{Only text}}                               & \multicolumn{2}{c|}{\textbf{Speech + text}}                             & \multicolumn{2}{c|}{\textbf{Only text}}                               \\ \cline{3-10} 
                                                                                                      & \multicolumn{1}{c|}{}             & \multicolumn{1}{c|}{\textbf{K-M}}  & \multicolumn{1}{c|}{\textbf{DNN}}  & \multicolumn{1}{c|}{\textbf{K-M}} & \multicolumn{1}{c|}{\textbf{DNN}} & \multicolumn{1}{c|}{\textbf{K-M}}  & \multicolumn{1}{c|}{\textbf{DNN}}  & \multicolumn{1}{c|}{\textbf{K-M}} & \multicolumn{1}{c|}{\textbf{DNN}} \\ \hline
\multicolumn{1}{|c|}{\multirow{6}{*}{\textbf{Native}}}                                                & \multicolumn{1}{c|}{\textbf{E}}   & \multicolumn{1}{c|}{\textbf{100}}  & \multicolumn{1}{c|}{\textbf{100}}  & \multicolumn{1}{c|}{75}           & \multicolumn{1}{c|}{\textbf{99}}  & \multicolumn{1}{c|}{--}            & \multicolumn{1}{c|}{--}            & \multicolumn{1}{c|}{--}           & \multicolumn{1}{c|}{--}           \\ \cline{2-10} 
\multicolumn{1}{|c|}{}                                                                                & \multicolumn{1}{c|}{\textbf{D}}   & \multicolumn{1}{c|}{84.9}          & \multicolumn{1}{c|}{92.5}          & \multicolumn{1}{c|}{70}           & \multicolumn{1}{c|}{89}           & \multicolumn{1}{c|}{--}            & \multicolumn{1}{c|}{--}            & \multicolumn{1}{c|}{--}           & \multicolumn{1}{c|}{--}           \\ \cline{2-10} 
\multicolumn{1}{|c|}{}                                                                                & \multicolumn{1}{c|}{\textbf{P}}   & \multicolumn{1}{c|}{80}            & \multicolumn{1}{c|}{85.8}          & \multicolumn{1}{c|}{68.5}         & \multicolumn{1}{c|}{84}           & \multicolumn{1}{c|}{--}            & \multicolumn{1}{c|}{--}            & \multicolumn{1}{c|}{--}           & \multicolumn{1}{c|}{--}           \\ \cline{2-10} 
\multicolumn{1}{|c|}{}                                                                                & \multicolumn{1}{c|}{\textbf{EDP}} & \multicolumn{1}{c|}{88.3}          & \multicolumn{1}{c|}{92.77}         & \multicolumn{1}{c|}{71.17}        & \multicolumn{1}{c|}{90.67}        & \multicolumn{1}{c|}{--}            & \multicolumn{1}{c|}{--}            & \multicolumn{1}{c|}{--}           & \multicolumn{1}{c|}{--}           \\ \cline{2-10} 
\multicolumn{1}{|c|}{}                                                                                & \multicolumn{1}{c|}{\textbf{HB}}  & \multicolumn{1}{c|}{66}            & \multicolumn{1}{c|}{86.5}          & \multicolumn{1}{c|}{--}           & \multicolumn{1}{c|}{--}           & \multicolumn{1}{c|}{--}            & \multicolumn{1}{c|}{--}            & \multicolumn{1}{c|}{--}           & \multicolumn{1}{c|}{--}           \\ \cline{2-10} 
\multicolumn{1}{|c|}{}                                                                                & \multicolumn{1}{c|}{\textbf{W2V}} & \multicolumn{1}{c|}{53.4}          & \multicolumn{1}{c|}{90.1}          & \multicolumn{1}{c|}{--}           & \multicolumn{1}{c|}{--}           & \multicolumn{1}{c|}{--}            & \multicolumn{1}{c|}{--}            & \multicolumn{1}{c|}{--}           & \multicolumn{1}{c|}{--}           \\ \hline
\multicolumn{1}{|c|}{\multirow{6}{*}{\textbf{\begin{tabular}[c]{@{}c@{}}Non-\\ Native\end{tabular}}}} & \multicolumn{1}{c|}{\textbf{E}}   & \multicolumn{1}{c|}{\textbf{84.2}} & \multicolumn{1}{c|}{\textbf{89.5}} & \multicolumn{1}{c|}{81}           & \multicolumn{1}{c|}{87.9}         & \multicolumn{1}{c|}{\textbf{82.6}} & \multicolumn{1}{c|}{\textbf{87.4}} & \multicolumn{1}{c|}{80.3}         & \multicolumn{1}{c|}{87.7}         \\ \cline{2-10} 
\multicolumn{1}{|c|}{}                                                                                & \multicolumn{1}{c|}{\textbf{D}}   & \multicolumn{1}{c|}{80.1}          & \multicolumn{1}{c|}{86.6}          & \multicolumn{1}{c|}{77.2}         & \multicolumn{1}{c|}{84}           & \multicolumn{1}{c|}{81.7}          & \multicolumn{1}{c|}{85}            & \multicolumn{1}{c|}{70.7}         & \multicolumn{1}{c|}{84.5}         \\ \cline{2-10} 
\multicolumn{1}{|c|}{}                                                                                & \multicolumn{1}{c|}{\textbf{P}}   & \multicolumn{1}{c|}{78}            & \multicolumn{1}{c|}{83.1}          & \multicolumn{1}{c|}{73.9}         & \multicolumn{1}{c|}{80.4}         & \multicolumn{1}{c|}{80.5}          & \multicolumn{1}{c|}{82.4}          & \multicolumn{1}{c|}{61.9}         & \multicolumn{1}{c|}{81.6}         \\ \cline{2-10} 
\multicolumn{1}{|c|}{}                                                                                & \multicolumn{1}{c|}{\textbf{EDP}} & \multicolumn{1}{c|}{80.8}          & \multicolumn{1}{c|}{86.4}          & \multicolumn{1}{c|}{77.4}         & \multicolumn{1}{c|}{84.1}         & \multicolumn{1}{c|}{81.6}          & \multicolumn{1}{c|}{84.9}          & \multicolumn{1}{c|}{71}           & \multicolumn{1}{c|}{84.6}         \\ \cline{2-10} 
\multicolumn{1}{|c|}{}                                                                                & \multicolumn{1}{c|}{\textbf{HB}}  & \multicolumn{1}{c|}{75.8}          & \multicolumn{1}{c|}{78.5}          & \multicolumn{1}{c|}{--}           & \multicolumn{1}{c|}{--}           & \multicolumn{1}{c|}{75.1}          & \multicolumn{1}{c|}{77.5}          & \multicolumn{1}{c|}{--}           & \multicolumn{1}{c|}{--}           \\ \cline{2-10} 
\multicolumn{1}{|c|}{}                                                                                & \multicolumn{1}{c|}{\textbf{W2V}} & \multicolumn{1}{c|}{56.8}          & \multicolumn{1}{c|}{82}            & \multicolumn{1}{c|}{--}           & \multicolumn{1}{c|}{--}           & \multicolumn{1}{c|}{55.5}          & \multicolumn{1}{c|}{81.8}          & \multicolumn{1}{c|}{--}           & \multicolumn{1}{c|}{--}           \\ \hline
                                                                                                      &                                   &                                    &                                    &                                   &                                   &                                    &                                    &                                   &                                   \\ \hline
\multicolumn{1}{|c|}{\multirow{6}{*}{\textbf{GER}}}                                                   & \multicolumn{1}{c|}{\textbf{E}}   & \multicolumn{1}{c|}{\textbf{92.2}} & \multicolumn{1}{c|}{\textbf{96.7}} & \multicolumn{1}{c|}{70}           & \multicolumn{1}{c|}{89.4}         & \multicolumn{1}{c|}{\textbf{88.6}} & \multicolumn{1}{c|}{\textbf{94.5}} & \multicolumn{1}{c|}{80.5}         & \multicolumn{1}{c|}{89.3}         \\ \cline{2-10} 
\multicolumn{1}{|c|}{}                                                                                & \multicolumn{1}{c|}{\textbf{D}}   & \multicolumn{1}{c|}{88}            & \multicolumn{1}{c|}{91.5}          & \multicolumn{1}{c|}{81}           & \multicolumn{1}{c|}{88.4}         & \multicolumn{1}{c|}{80.2}          & \multicolumn{1}{c|}{91.6}          & \multicolumn{1}{c|}{78.5}         & \multicolumn{1}{c|}{88.3}         \\ \cline{2-10} 
\multicolumn{1}{|c|}{}                                                                                & \multicolumn{1}{c|}{\textbf{P}}   & \multicolumn{1}{c|}{84.2}          & \multicolumn{1}{c|}{90}            & \multicolumn{1}{c|}{77.3}         & \multicolumn{1}{c|}{71.6}         & \multicolumn{1}{c|}{72.5}          & \multicolumn{1}{c|}{88}            & \multicolumn{1}{c|}{70.3}         & \multicolumn{1}{c|}{73.6}         \\ \cline{2-10} 
\multicolumn{1}{|c|}{}                                                                                & \multicolumn{1}{c|}{\textbf{EDP}} & \multicolumn{1}{c|}{88.1}          & \multicolumn{1}{c|}{95.5}          & \multicolumn{1}{c|}{76.1}         & \multicolumn{1}{c|}{83.1}         & \multicolumn{1}{c|}{80.4}          & \multicolumn{1}{c|}{93.4}          & \multicolumn{1}{c|}{76.4}         & \multicolumn{1}{c|}{83.7}         \\ \cline{2-10} 
\multicolumn{1}{|c|}{}                                                                                & \multicolumn{1}{c|}{\textbf{HB}}  & \multicolumn{1}{c|}{67}            & \multicolumn{1}{c|}{83}            & \multicolumn{1}{c|}{--}           & \multicolumn{1}{c|}{--}           & \multicolumn{1}{c|}{62.6}          & \multicolumn{1}{c|}{88.6}          & \multicolumn{1}{c|}{--}           & \multicolumn{1}{c|}{--}           \\ \cline{2-10} 
\multicolumn{1}{|c|}{}                                                                                & \multicolumn{1}{c|}{\textbf{W2V}} & \multicolumn{1}{c|}{63.5}          & \multicolumn{1}{c|}{80.5}          & \multicolumn{1}{c|}{--}           & \multicolumn{1}{c|}{--}           & \multicolumn{1}{c|}{57.7}          & \multicolumn{1}{c|}{87.6}          & \multicolumn{1}{c|}{--}           & \multicolumn{1}{c|}{--}           \\ \hline
\multicolumn{1}{|c|}{\multirow{6}{*}{\textbf{ITA}}}                                                   & \multicolumn{1}{c|}{\textbf{E}}   & \multicolumn{1}{c|}{\textbf{90}}   & \multicolumn{1}{c|}{\textbf{92.3}} & \multicolumn{1}{c|}{74}           & \multicolumn{1}{c|}{75.4}         & \multicolumn{1}{c|}{\textbf{81.7}} & \multicolumn{1}{c|}{\textbf{90.5}} & \multicolumn{1}{c|}{78.2}         & \multicolumn{1}{c|}{80.3}         \\ \cline{2-10} 
\multicolumn{1}{|c|}{}                                                                                & \multicolumn{1}{c|}{\textbf{D}}   & \multicolumn{1}{c|}{81}            & \multicolumn{1}{c|}{87.7}          & \multicolumn{1}{c|}{71}           & \multicolumn{1}{c|}{73.7}         & \multicolumn{1}{c|}{77.9}          & \multicolumn{1}{c|}{81.8}          & \multicolumn{1}{c|}{73.9}         & \multicolumn{1}{c|}{75.1}         \\ \cline{2-10} 
\multicolumn{1}{|c|}{}                                                                                & \multicolumn{1}{c|}{\textbf{P}}   & \multicolumn{1}{c|}{76}            & \multicolumn{1}{c|}{84.7}          & \multicolumn{1}{c|}{68.5}         & \multicolumn{1}{c|}{72.8}         & \multicolumn{1}{c|}{73.4}          & \multicolumn{1}{c|}{83.9}          & \multicolumn{1}{c|}{72.2}         & \multicolumn{1}{c|}{71.4}         \\ \cline{2-10} 
\multicolumn{1}{|c|}{}                                                                                & \multicolumn{1}{c|}{\textbf{EDP}} & \multicolumn{1}{c|}{82.3}          & \multicolumn{1}{c|}{88.2}          & \multicolumn{1}{c|}{71.2}         & \multicolumn{1}{c|}{74}           & \multicolumn{1}{c|}{77.7}          & \multicolumn{1}{c|}{84.5}          & \multicolumn{1}{c|}{74.8}         & \multicolumn{1}{c|}{75.6}         \\ \cline{2-10} 
\multicolumn{1}{|c|}{}                                                                                & \multicolumn{1}{c|}{\textbf{HB}}  & \multicolumn{1}{c|}{74}            & \multicolumn{1}{c|}{79}            & \multicolumn{1}{c|}{--}           & \multicolumn{1}{c|}{--}           & \multicolumn{1}{c|}{57.7}          & \multicolumn{1}{c|}{88.2}          & \multicolumn{1}{c|}{--}           & \multicolumn{1}{c|}{--}           \\ \cline{2-10} 
\multicolumn{1}{|c|}{}                                                                                & \multicolumn{1}{c|}{\textbf{W2V}} & \multicolumn{1}{c|}{50.3}          & \multicolumn{1}{c|}{83.3}          & \multicolumn{1}{c|}{--}           & \multicolumn{1}{c|}{--}           & \multicolumn{1}{c|}{56.6}          & \multicolumn{1}{c|}{89.3}          & \multicolumn{1}{c|}{--}           & \multicolumn{1}{c|}{--}           \\ \hline
\end{tabular}
\label{table:Accuracies}
\vspace{-0.6cm}
\end{table}

The top half of the table shows the accuracies for native and non-native scenarios, whereas the bottom half presents detailed non-native specific analysis obtained for German (GER) and Italian (ITA) separately. The accuracies from the top half indicate that the embeddings yield higher accuracy in native scenarios. Notably, DNN and K-Means gave 100\% accuracy with energy embeddings in native ``speech plus text" case because of the complete separation of stressed and unstressed components as observed in the PCA analysis, indicating no overlap.  However, the accuracies under non-native speech are lesser than those under the native speech for both ``speech plus text" and ``text-only" cases. Further, the accuracies under non-native speech are higher for ``speech plus text" case compared to ``text-only" case. These two observations together suggest that embeddings exhibit better representation in the presence of ``speech plus text" for non-native cases also, emphasizing sensitivity to nuances in syllabic emphasis.\par From the bottom half of the table, GER consistently performs well in both ``speech plus text" and ``text-only", while ITA excels mainly in ``speech plus text" for word prominence and a similar trend is observed for syllable prominence between GER and ITA. This could be due to the linguistic proximity between English and German, both Germanic languages, contributes to their higher similarity compared to English and Romance languages (French, Italian, Spanish), which originate from Latin \cite{8076005}. Therefore, the FastSpeech2 model, pretrained on the native LJSpeech Corpus, demonstrates better performance on non-native English spoken by GER than ITA.\par

\begin{figure}[ht]
  \centering
  \includegraphics [width=0.5\textwidth] {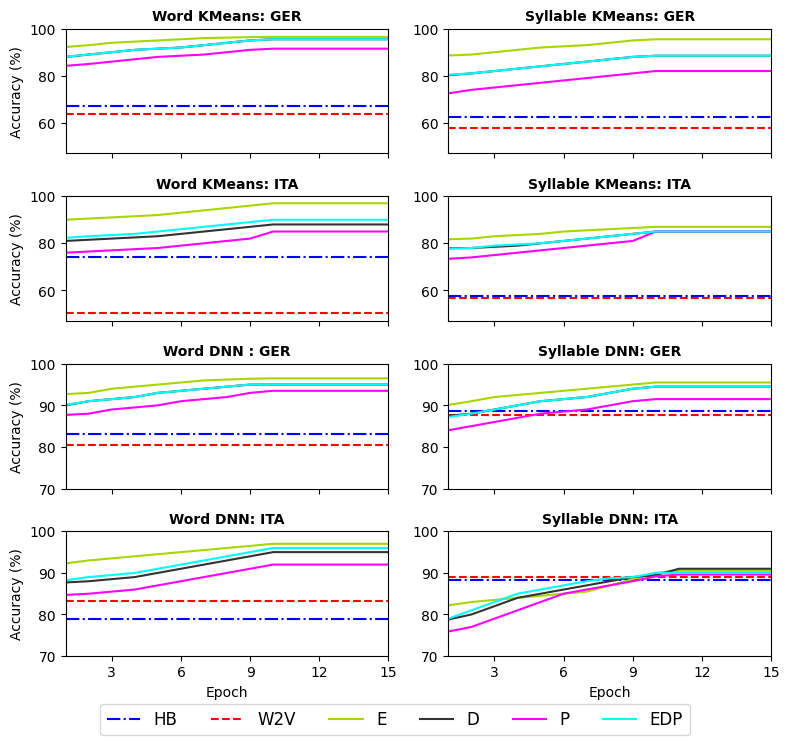}
  \vspace{-0.65cm}  
  \caption{Comparison of Epoch-wise Accuracies with K-Means and DNN classification for GER and ITA considering word-level and syllable-level prominence detection.}
  \label{fig:epoch_acc}
  \vspace{-0.65cm}
\end{figure}
 The Figure \ref{fig:epoch_acc} illustrates the epoch-wise accuracies of syllable and word-level prominence detection for non-native speech, comparing K-Means and DNN against heuristics and Wav2Vec-2.0 representations. As the number of epochs increases, the accuracy for both GER and ITA shows a clear improvement, eventually reaching a point of gradual saturation. Notably, the accuracy not only enhances with extended training but also consistently surpasses the baseline accuracy established by heuristics and Wav2Vec-2.0 representations. This indicates that the embeddings with TTS training mode with multiple epochs could capture the prominence patterns in non-native speech very well, leading to superior performance compared to traditional methods.

\vspace{-0.12cm}
\section{Conclusion}
\vspace{-0.1cm}

The study conducted in this work infers that the energy embeddings, particularly those obtained during native ``speech plus text" case, consistently outperform other scenarios. They also exhibit superior performance compared to duration and pitch embeddings in the native as well as non-native ``text-only" case. Distances computed to measure the separation between prosody embeddings corresponding to stressed and unstressed words and syllables indicate minimal similarity and a notable degree of dissimilarity. This observation holds true in clustering and classification experiments across various scenarios also, except for specifically Italian speakers where embeddings do not surpass baseline accuracies. However, with increasing epochs, the ``speech plus text" embeddings for Italian also exceed the accuracies achieved by other baselines considered in this work. This deviation can be attributed to the TTS pretraining on a native English dataset, with German exhibiting phonological similarities to English, contributing to enhanced performance. 


\bibliographystyle{IEEEbib}
\bibliography{refs}

\begin{thebibliography}{10}

\bibitem{werner}
Stefan Werner and Eric Keller,
\newblock {\em Prosodic aspects of speech},
\newblock 1995.

\bibitem{lehiste1970suprasegmentals}
Ilse Lehiste,
\newblock {\em Suprasegmentals},
\newblock MIT Press, Cambridge, 1970.

\bibitem{cutler2005lexical}
Anne Cutler,
\newblock ``Lexical stress,''
\newblock in {\em The Handbook of Speech Perception}. Oxford, 2005.

\bibitem{wp1}
Dagen Wang and S.~Narayanan,
\newblock ``An unsupervised quantitative measure for word prominence in spontaneous speech,''
\newblock in {\em Proceedings. (ICASSP '05). IEEE International Conference on Acoustics, Speech, and Signal Processing, 2005.}

\bibitem{wp3}
Taniya Mishra, Vivek Kumar~Rangarajan Sridhar, and Alistair Conkie,
\newblock ``Word prominence detection using robust yet simple prosodic features,''
\newblock in {\em Interspeech}, 2012.

\bibitem{wp2}
Dagen Wang and Shrikanth Narayanan,
\newblock ``An acoustic measure for word prominence in spontaneous speech,''
\newblock {\em IEEE Transactions on Audio, Speech, and Language Processing}, vol. 15, no. 2, pp. 690--701, 2007.

\bibitem{Tepperman}
J.~Tepperman and S.~Narayanan,
\newblock ``Automatic syllable stress detection using prosodic features for pronunciation evaluation of language learners,''
\newblock in {\em Proceedings. (ICASSP '05). IEEE International Conference on Acoustics, Speech, and Signal Processing, 2005.}

\bibitem{yarra19b_slate}
Chiranjeevi Yarra, Manoj~Kumar Ramanathi, and Prasanta~Kumar Ghosh,
\newblock ``{Comparison of automatic syllable stress detection quality with time-aligned boundaries and context dependencies},''
\newblock in {\em Proc. 8th ISCA Workshop on Speech and Language Technology in Education (SLaTE 2019)}.

\bibitem{7953277}
Chiranjeevi Yarra, Om~D. Deshmukh, and Prasanta~Kumar Ghosh,
\newblock ``Automatic detection of syllable stress using sonority based prominence features for pronunciation evaluation,''
\newblock in {\em 2017 IEEE International Conference on Acoustics, Speech and Signal Processing (ICASSP)}.

\bibitem{vae-dnn}
Jhansi Mallela, Prasanth~Sai Boyina, and Chiranjeevi Yarra,
\newblock ``A comparison of learned representations with jointly optimized vae and dnn for syllable stress detection,''
\newblock in {\em Speech and Computer: 25th International Conference, SPECOM}, 2023.

\bibitem{NCC_W2VEC}
Jhansi Mallela, Sai~Harshitha Aluru, and Chiranjeevi Yarra,
\newblock ``Exploring the use of self-supervised representations for automatic syllable stress detection,''
\newblock in {\em 2024 National Conference on Communications (NCC)}.

\bibitem{Tacotron2}
Jonathan Shen, Ruoming Pang, Ron~J. Weiss, et~al.,
\newblock ``Natural tts synthesis by conditioning wavenet on mel spectrogram predictions,''
\newblock in {\em 2018 IEEE International Conference on Acoustics, Speech and Signal Processing (ICASSP)}.

\bibitem{Tacotron}
Yuxuan Wang, R.~J. Skerry-Ryan, et~al.,
\newblock ``Tacotron: Towards end-to-end speech synthesis,''
\newblock in {\em Interspeech}, 2017.

\bibitem{Glow-TTS}
Jaehyeon Kim, Sungwon Kim, et~al.,
\newblock ``Glow-tts: A generative flow for text-to-speech via monotonic alignment search,''
\newblock in {\em Advances in Neural Information Processing Systems}, 2020, vol.~33.

\bibitem{Glow}
Diederik~P. Kingma and Prafulla Dhariwal,
\newblock ``Glow: generative flow with invertible 1\texttimes{}1 convolutions,''
\newblock in {\em Proceedings of the 32nd International Conference on Neural Information Processing Systems}, 2018.

\bibitem{vits}
Jaehyeon Kim, Jungil Kong, and Juhee Son,
\newblock ``Conditional variational autoencoder with adversarial learning for end-to-end text-to-speech,''
\newblock in {\em Proceedings of the 38th International Conference on Machine Learning}, 2021.

\bibitem{flow}
Danilo~Jimenez Rezende and Shakir Mohamed,
\newblock ``Variational inference with normalizing flows,''
\newblock in {\em Proceedings of the 32nd International Conference on on Machine Learning}, 2015.

\bibitem{ren2022fastspeech}
Yi~Ren, Chenxu Hu, et~al.,
\newblock ``Fastspeech 2: Fast and high-quality end-to-end text to speech,''
\newblock in {\em International Conference on Learning Representations}, 2021.

\bibitem{FastSpeech}
Yi~Ren, Yangjun Ruan, et~al.,
\newblock {\em FastSpeech: fast, robust and controllable text to speech},
\newblock Proceedings of the 33rd International Conference on Neural Information Processing Systems, 2019.

\bibitem{tatoeba_url}
``Tatoeba,'' \url{https://tatoeba.org/en}.

\bibitem{ISLE}
ES~Atwell, PA~Howarth, and DC~Souter,
\newblock ``The isle corpus: Italian and german spoken learner's english,''
\newblock {\em ICAME Journal: International Computer Archive of Modern and Medieval English Journal}, 2003.

\bibitem{tatoeba}
Artetxe Mikel and Schwenk Holger,
\newblock ``Massively multilingual sentence embeddings for zero-shot cross-lingual transfer and beyond,''
\newblock {\em arXiv:1812.10464v2}, 2018.

\bibitem{Tiedemann2012ParallelDT}
J{\"o}rg Tiedemann,
\newblock ``Parallel data, tools and interfaces in opus,''
\newblock in {\em International Conference on Language Resources and Evaluation}, 2012.

\bibitem{NIST}
B.~Fisher,
\newblock ``tsylb2-1.1 syllabification software, national institute of standards and technology,'' 1996.

\bibitem{ljspeech17}
Keith Ito and Linda Johnson,
\newblock ``The lj speech dataset,'' \url{https://keithito.com/LJ-Speech-Dataset/}, 2017.

\bibitem{wav2vec}
Alexei Baevski, Henry Zhou, et~al.,
\newblock ``wav2vec 2.0: a framework for self-supervised learning of speech representations,''
\newblock in {\em Proceedings of the 34th International Conference on Neural Information Processing Systems}, 2020, NIPS '20.

\bibitem{distance}
Vivek Mehta et~al.,
\newblock ``Analytical review of clustering techniques and proximity measures,''
\newblock {\em Artificial Intelligence Review}, 2020.

\bibitem{8076005}
Lütfi~Kerem Şenel, Veysel Yücesoy, et~al.,
\newblock ``Measuring cross-lingual semantic similarity across european languages,''
\newblock in {\em 2017 40th International Conference on Telecommunications and Signal Processing (TSP)}.

\end{thebibliography}
\end{document}